\documentclass{article}
\usepackage{styles/acra}
\usepackage{acronym}
\usepackage{amsmath}
\usepackage{amssymb}
\usepackage{url}
\usepackage{xcolor}
\usepackage{fancyhdr}

\usepackage{adjustbox}
\usepackage{graphicx}
\usepackage{tensor}
\usepackage[utf8]{inputenc}
\usepackage[OT1]{fontenc}
\usepackage[final]{microtype}
\usepackage{booktabs}
\usepackage[binary-units]{siunitx}
\usepackage{multirow}
\usepackage{subcaption}
\usepackage{lipsum}

\makeatletter
\let\NAT@parse\undefined
\makeatother
\usepackage[bookmarks=true, colorlinks=true, citecolor=blue, linkcolor=blue]{hyperref}
\usepackage[capitalise]{cleveref}

\usepackage[numbers]{natbib}

\acrodef{LiDAR}{Light Detection and Ranging}
\acrodef{IMU}{Inertial Measurement Unit}
\acrodef{BEV}{Bird's-Eye View}
\acrodef{NN}{Neural Network}
\acrodef{FoV}{Field of View}
\acrodef{PQ}{Panoptic Quality}
\acrodef{SQ}{Segmentation Quality}
\acrodef{RQ}{Recognition Quality}

\title{\LARGE \bf
ETHcavation: A Dataset and Pipeline for Panoptic Scene Understanding and Object Tracking in Dynamic Construction Environments}

\author{
\begin{minipage}[t]{\textwidth}
    \centering
    Lorenzo Terenzi$^\dagger$, Julian Nubert$^{\dagger, \ddagger}$, Pol Eyschen$^\dagger$, Pascal Roth$^\dagger$, \\
    Simin Fei$^\dagger$, Edo Jelavic$^\dagger$, and Marco Hutter$^\dagger$\thanks{This work is supported by the NCCR digital fabrication \& robotics, the SNF project No. 188596, and the Max Planck ETH Center for Learning Systems.}
\end{minipage} \\
$^\dagger$ Robotic Systems Lab, ETH Z\"urich, $^\ddagger$ MPI for Intelligent Systems, Stuttgart, Germany \\
Corresponding Author: Lorenzo Terenzi, \href{mailto:lterenzi@ethz.ch}{lterenzi@ethz.ch}
}

\begin{document}

\maketitle
\thispagestyle{empty}
\pagestyle{empty}

%%%%%%%%%%%%%%%%%%%%%%%%%%%%%%%%%%%%%%%%%%%%%%%%%%%%%%%%%%%%%%%%%%%%%%%%%%%%%%%%
\begin{abstract}
Construction sites are challenging environments for autonomous systems due to their unstructured nature and the presence of dynamic actors, such as workers and machinery. This work presents a comprehensive panoptic scene understanding solution designed to handle the complexities of such environments by integrating 2D panoptic segmentation with 3D LiDAR mapping. Our system generates detailed environmental representations in real-time by combining semantic and geometric data, supported by Kalman Filter-based tracking for dynamic object detection.
We introduce a fine-tuning method that adapts large pre-trained panoptic segmentation models for construction site applications using a limited number of domain-specific samples. For this use case, we release a first-of-its-kind dataset of 502 hand-labeled sample images with panoptic annotations from construction sites. In addition, we propose a dynamic panoptic mapping technique that enhances scene understanding in unstructured environments. As a case study, we demonstrate the system's application for autonomous navigation, utilizing real-time RRT* for reactive path planning in dynamic scenarios.
The dataset\footnote{\label{footnote:project_page}\scriptsize\url{https://leggedrobotics.github.io/panoptic-scene-understanding.github.io/}} and code\footnote{\label{footnote:code}\scriptsize\url{https://github.com/leggedrobotics/rsl_panoptic_mapping}} for training and deployment are publicly available to support future research.
\end{abstract}

\begin{figure}[!htb]
    \centering
    % First image
    \includegraphics[width=\linewidth, clip, trim=0px 300px 0px 300px]{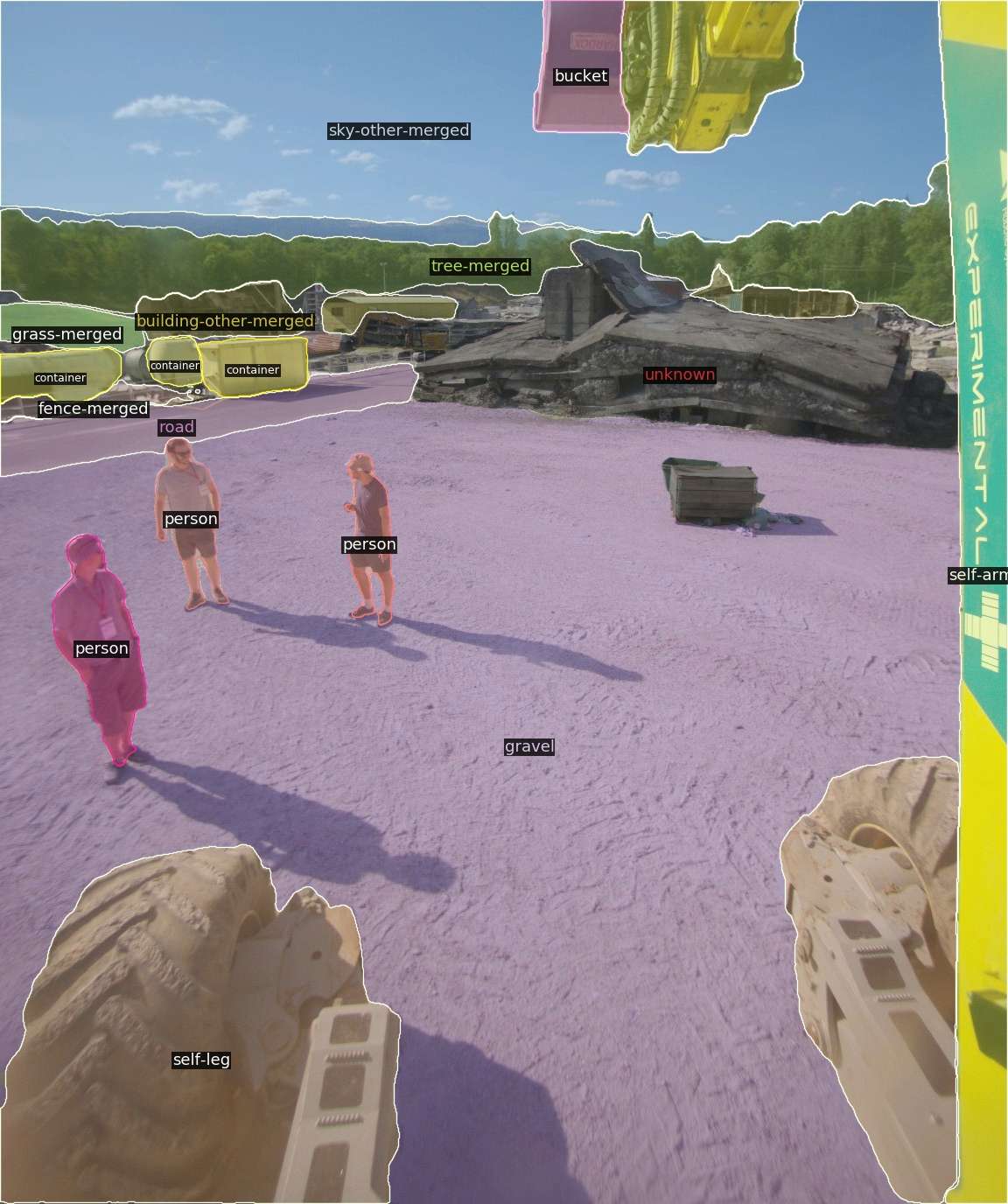}
    %\caption{Optional caption for the first image}
    \label{fig:top_image}
    
    % Remove any spacing between the images if needed
    \vspace{-10pt} % Adjust the negative space to remove any unwanted space between the images

    % Second image, directly below the first one
    \includegraphics[width=\linewidth, clip, trim=0px 0px 0px 100px]{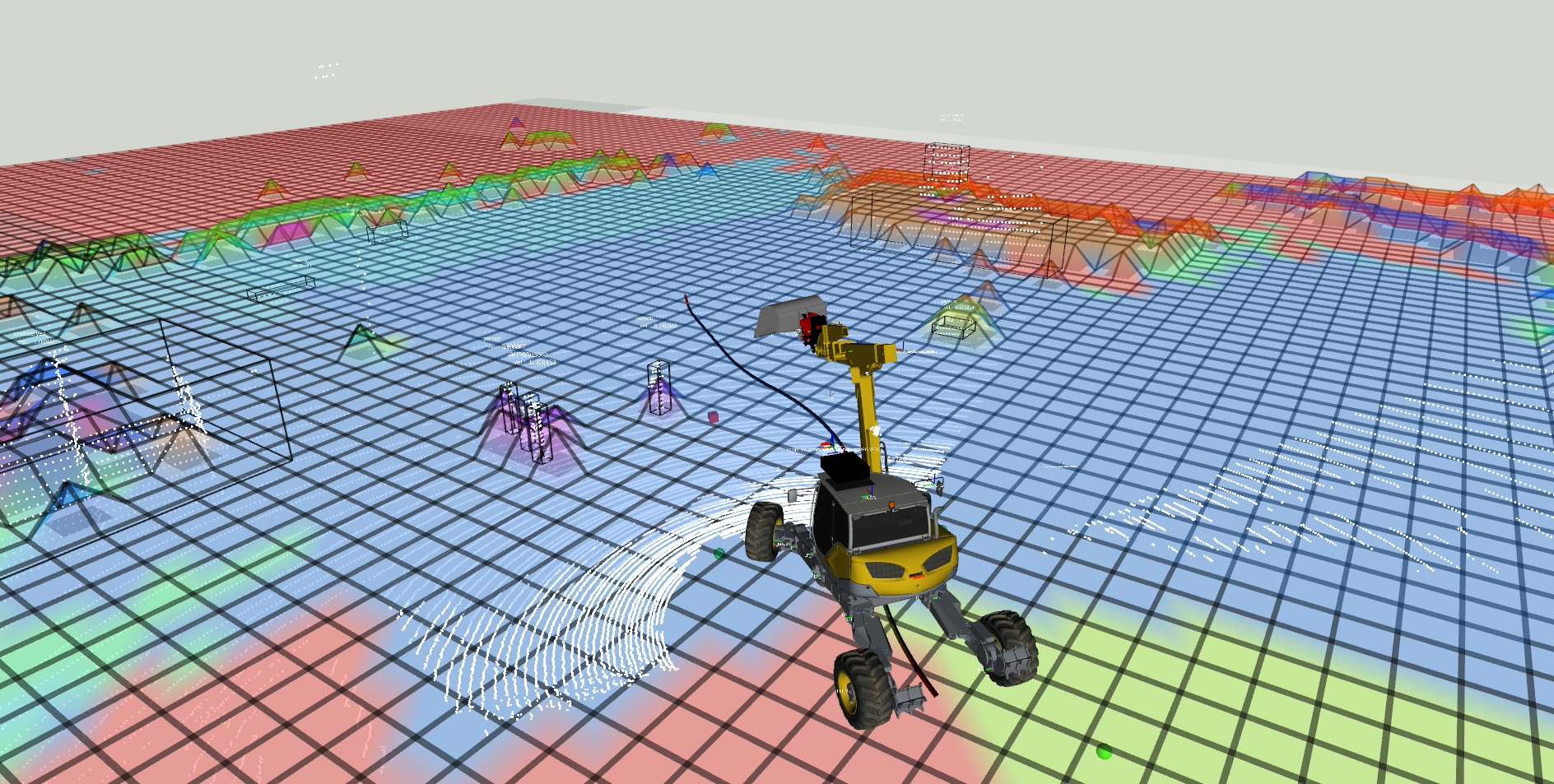}
    %\caption{Optional caption for the second image}
    \label{fig:bottom_image}

    \caption{Autonomous navigation with the \textit{M545} excavator. The top image illustrates the model's 2D panoptic segmentation prediction, while the bottom image depicts the panoptic map and the planned path of the navigation planner in black.}
    \label{fig:title}
\end{figure}

\section{Introduction}\label{sec:intro}

The construction industry is experiencing a growing need for advanced perception systems to tackle the challenges of dynamic, cluttered, and unstructured environments. As autonomous machinery becomes a focal point for improving safety, efficiency, and productivity in construction, robust scene understanding systems are crucial. Panoptic segmentation, which combines semantic and instance segmentation, offers a promising solution by providing pixel-level semantic labels and object-level instance boundaries. When integrated with 3D \ac{LiDAR} data, panoptic segmentation enables more detailed and accurate 3D environmental models, essential for autonomous decision-making in construction sites.

However, developing effective panoptic segmentation models for construction environments is challenging. Real-world construction sites present unique obstacles, such as varying terrain, dynamic actors (e.g., workers and machinery), and limited availability of labeled data to train perception networks. Previous efforts, like those by Guan et al.~\cite{guan2022tns}, have shown that specialized construction datasets can improve segmentation performance, but noise and limited class diversity remain. Furthermore, integrating \ac{LiDAR} data - while enhancing geometric understanding - adds complexity to training and deployment due to the difficulty of annotating 3D data in such environments. 

This work addresses these challenges by developing a comprehensive panoptic segmentation system tailored explicitly for dynamic construction sites. Our approach fine-tunes pre-trained panoptic segmentation models using a custom small construction dataset to improve generalization with limited training data. We also propose a dynamic panoptic mapping technique that fuses 2D image-based segmentation with 3D \ac{LiDAR} data to create detailed, real-time maps of construction environments. These maps are enriched with semantic and geometric information, providing a robust foundation for downstream applications, such as navigation.

As a case study, we demonstrate the application of our panoptic segmentation system for autonomous navigation in dynamic construction sites, as shown in \cref{fig:title}. Leveraging the generated dynamic panoptic maps, we integrate them into an online RRT* planner, enabling reactive path planning in unstructured environments. The top image in \cref{fig:title} illustrates the model's panoptic segmentation prediction of the scene. In contrast, the bottom image depicts the corresponding panoptic map and the resulting planned path in black. 

Our contributions include:
\begin{enumerate}
    \item We present a structured approach for fine-tuning large, pre-trained panoptic segmentation models on specialized datasets. We provide recommendations on training regimes, model sizes, and dataset sizes.
    \item We propose a method for creating dynamic panoptic maps in environments with scarce labeled data. This method combines image panoptic segmentation and \ac{LiDAR}-based object detection and tracking to enable robust and safe navigation planning.
    \item We introduce a dataset of 502 images with hand-labeled panoptic annotations from multiple locations covering 35 actively selected semantic categories.\textsuperscript{\ref{footnote:project_page}}
    \item We release the code for model fine-tuning and semantic mapping to foster future robotic research.\textsuperscript{\ref{footnote:code}}
\end{enumerate}

%%%%%%%%%%%%%%%%%%%%%%%%%%%%%%%%%%%%%%%%%%%%%%%%%%%%%%%%%%%%%%%%%%%%%%%%%%%%%%%%

%%%%%%%%%%%%%%%%%%%%%%%%%%%%%%%%%%%%%%%%%%%%%%%%%%%%%%%%%%%%%%%%%%%%%%%%%%%%%%%%
\section{Related Work}\label{sec:related}

The following sections summarize recent advancements in semantic scene understanding, focusing on transformer-based panoptic segmentation and its implications for motion planning.

\subsection{Semantic \& Panoptic Segmentation}
The segmentation field has witnessed significant progress with the advent of semantic and instance segmentation techniques, particularly with the introduction of transformer-based~\cite{vaswani2017attention} architectures for panoptic segmentation. Semantic segmentation assigns a class to each pixel, such as \textit{'sky'} or \textit{'road'}, whereas instance segmentation identifies and delineates each object instance. Pioneering methods such as Fully Convolutional Networks (FCN)~\cite{long2015fully} and DeepLab~\cite{liang2017rethinking} laid the groundwork for contemporary segmentation models. Following this, instance segmentation was revolutionized by frameworks like Mask R-CNN~\cite{he2017maskrcnn}.

Panoptic segmentation~\cite{kirillov2019panoptic} introduced a unified approach to segmentation, merging semantic and instance segmentation tasks. This led to the creation of models such as Panoptic FPN~\cite{kirillov2019panopticfpn} and EfficientPS~\cite{mohan2020efficientps}, incorporating features from both segmentation domains.

A notable leap in segmentation quality was achieved through transformer-based decoders, with models like DETR~\cite{carion2020end} setting new benchmarks at the time. Employing pre-trained transformer encoders, such as Swin and ViT, trained in a supervised manner on large-scale datasets like ImageNet~\cite{deng2009imagenet}, has substantially enhanced panoptic segmentation results over the past few years. 
More recent models, including MaskFormer~\cite{cheng2021maskformer}, Mask2Former~\cite{cheng2021mask2former}, and PanopticSegformer~\cite{li2021segformer} exemplify this advancement, offering significant improvements in model output quality.
Motivated by this, we investigate the usage and fine-tuning of DETR and Mask2Former in this work and highlight that Mask2Former can significantly outperform older architectures like DETR.

\subsection{Motion Planning with Semantic Understanding}
Panoptic segmentation enhances motion planning in autonomous driving by allowing a more fine-grained understanding of the surrounding environment.
In 3D object detection, \acp{LiDAR} and visible light cameras are the most common sensors. Among the \ac{LiDAR}-based detection state-of-the-art systems are pillar-based networks~\cite{liPillarNeXtRethinkingNetwork2023}, sparse 3D voxel transformers~\cite{wangDSVTDynamicSparse2023}, and PointNet-like transformers~\cite{wuPointTransformerV32023}. Conversely, camera-only approaches, such as those converting 2D images to bird's-eye view maps for planning~\cite{philionLiftSplatShoot2020,roddickPredictingSemanticMap2020}, offer cheaper alternative solutions but generally lag in geometric precision when compared to \ac{LiDAR}-based methods. These camera-only approaches typically project camera images into frustum-shaped features or employ dense transform layers to generate semantic occupancy maps from multi-scale image features.

Hybrid methods that integrate both \ac{LiDAR} and camera data often propose end-to-end trained \ac{NN} solutions that rely on extensive datasets, such as Nuscenes~\cite{caesar2020nuscenes} and Waymo Open Dataset~\cite{mei2022waymo}, which include labeled point cloud and image data.
Examples are~\cite{liuBEVFusionMultiTaskMultiSensor2022b,hu2023ealss,frey2024roadrunner}, producing \ac{BEV} representations for planning and demonstrating slightly superior performance in 3D object detection compared to \ac{LiDAR}-only solutions. Moreover, using imperative learning, Roth et al.~\cite{roth2023viplanner} learn a reactive local end-to-end navigation solution for urban navigation from depth-camera data and semantic images without explicitly creating a \ac{BEV} representation.

Challenges arise in domains lacking extensive labeled data, where neither autonomous driving datasets nor conventional methods generalize well. For \ac{LiDAR}-only applications, such as off-road navigation, sparse 3D Convolutional Neural Networks have been explored for classifying \ac{LiDAR} inputs into traversability classes~\cite{shabanSemanticTerrainClassification}. However, these methods struggle with a broad array of classes and terrains. Camera-only and hybrid sensor approaches face limitations in accurately identifying and classifying diverse objects and terrain types due to the inherent restrictions of range sensors and the complexity of integrating sensor data.

To address these challenges, recent efforts have focused on generating high-definition maps~\cite{jelavic2022open3d} from combined 2D images and 3D point clouds. This leads to coherent semantic maps projected onto occupancy grids, as discussed in~\cite{paz2020probabilistic}. Further, integrating semantic and geometric data for off-road navigation~\cite{maturana2018realtime,guan2022tns}, utilizing fused \ac{LiDAR} and camera information to create semantic 2D maps. These maps inform motion planning with predefined traversability costs but lack explicit handling of dynamic obstacles, occlusions, and entities outside the camera's field of view.

In this work, we also deploy an explicit mapping-based approach, where 2D panoptic image estimates are lifted to 3D through explicit lidar-based geometric projection. Moreover, we track objects through time and dynamically update the underlying map representation without forgetting, allowing for reliable operation in dynamic and complex environments.

\section{Proposed Approach}\label{sec:method}

%%%%%%%%%%%%%%%%%%%%%%%%%%%%%%%%%%%%%%%%%%%%%%%%%%%%%%%%%%%%%%%%%
\begin{figure}[t]
  \centering
  \includegraphics[width=\linewidth]{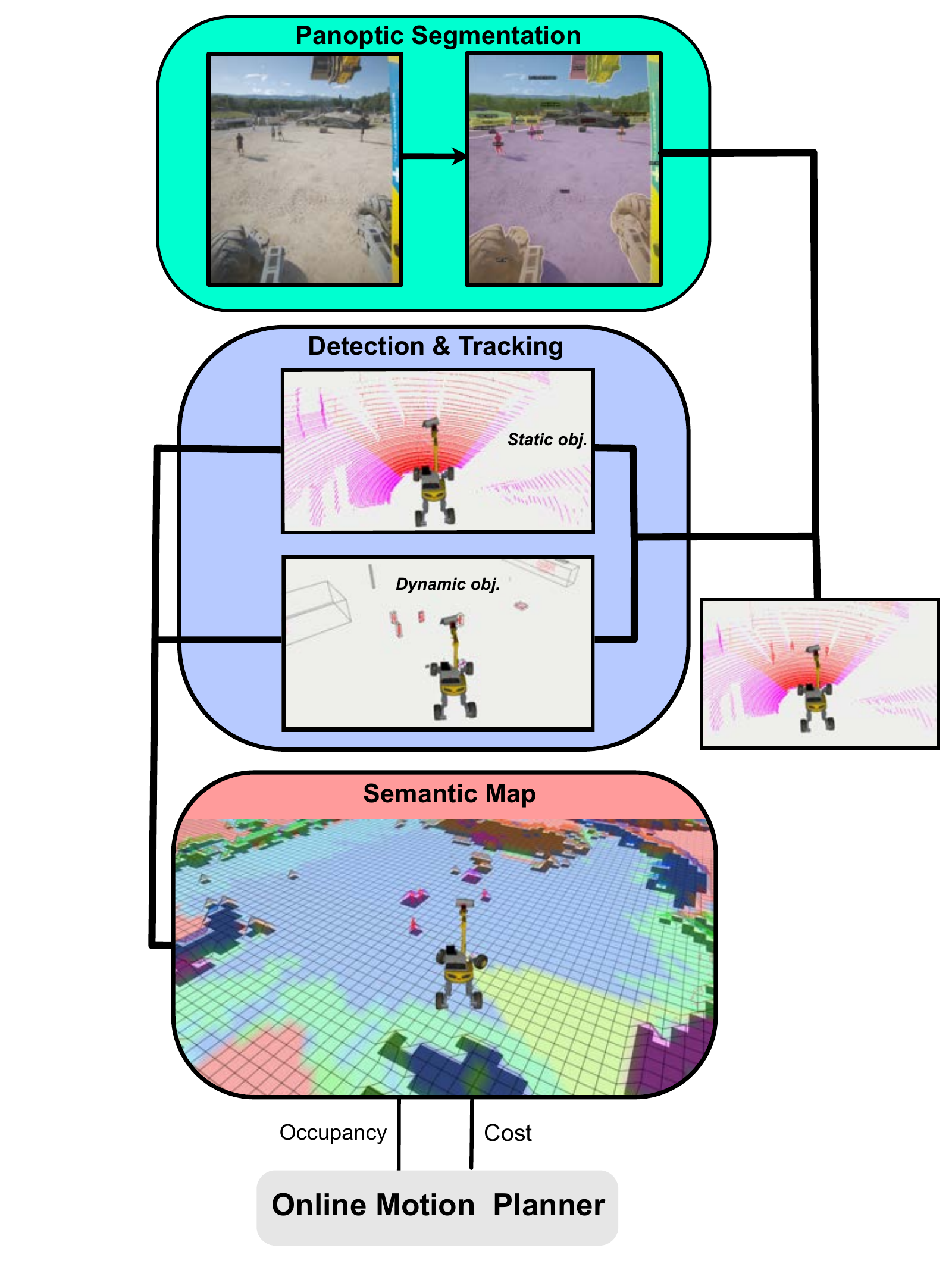} % Replace with your figure
  \caption{Overview of the integrated camera and \ac{LiDAR} data processing and semantic mapping pipeline for autonomous navigation in unstructured environments.}
  \label{fig:system_pipeline}
\end{figure}
%%%%%%%%%%%%%%%%%%%%%%%%%%%%%%%%%%%%%%%%%%%%%%%%%%%%%%%%%%%%%%%%%

This section outlines our method for combining camera and \ac{LiDAR} data to enable excavator navigation in unstructured settings. 

\subsubsection{System Overview}
We use a panoptic segmentation network (\cref{subsec:approach_panoptic_segmentation}) to process the camera feed, identify objects, and integrate this information into a 3D point cloud map via our mapping and tracking system (\cref{subsection:dynamic_mapping}). This allows for real-time map updates and object tracking. Utilizing this map, our navigation system creates a 2D traversability and cost map (\cref{subsubsec:semantic_map}). This is an ideal representation for real-world downstream tasks such as navigation (\cref{subsec:approach_navigation}). An overview of the system is presented in \cref{fig:system_pipeline}.

%%%%%%%%%%%%%%%%%%%%%%%%%%%%%%%%%%%%%%%%%%%%%%%%%%%%%%%%%%%%%%%%%
\begin{figure*}[t]
    \centering
    \begin{subfigure}{0.19\textwidth}
        \includegraphics[width=\linewidth]{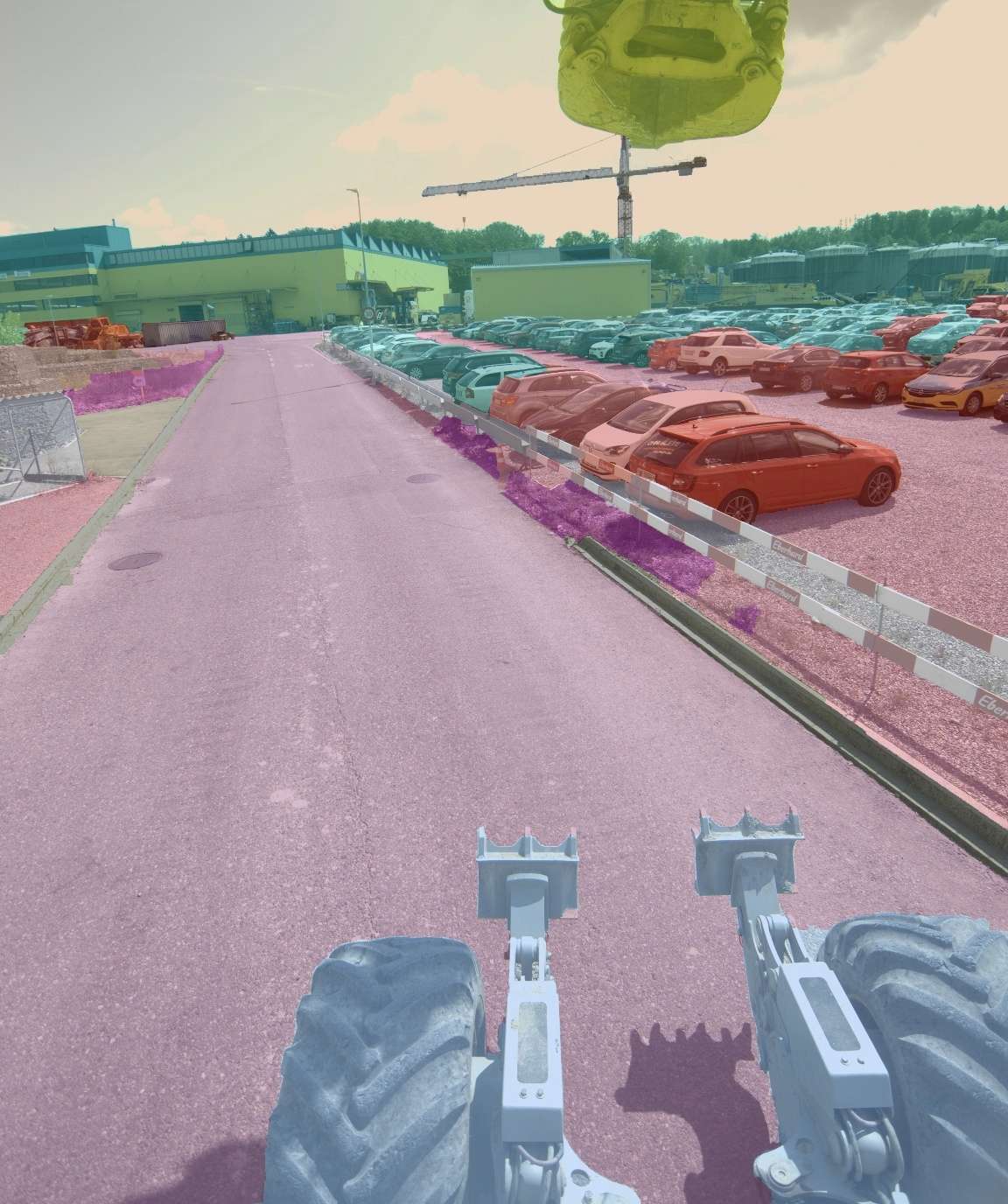} % Example image
        \label{fig:sub1}
    \end{subfigure}
    \hfill
    \begin{subfigure}{0.19\textwidth}
        \includegraphics[width=\linewidth]{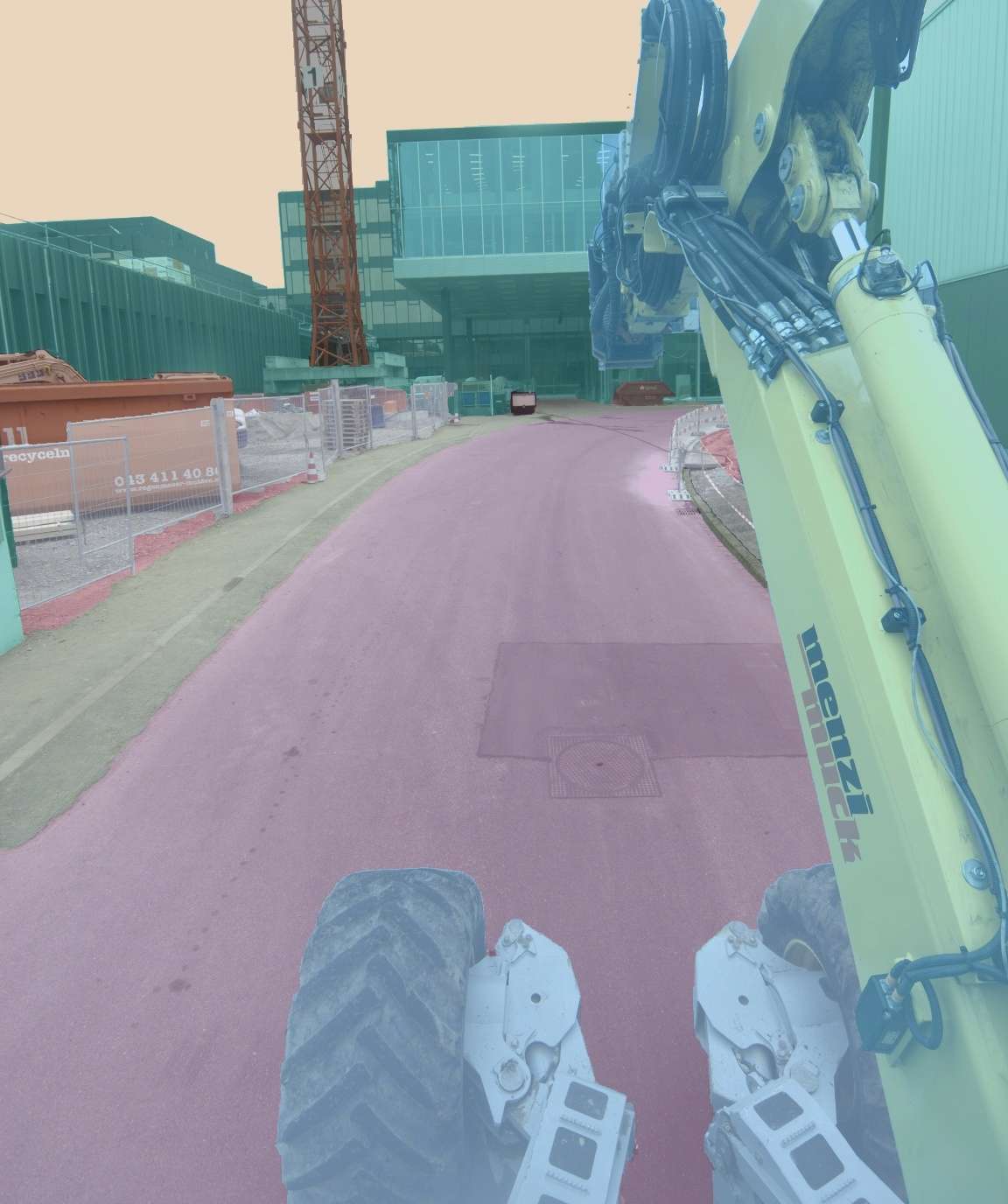}
        \label{fig:sub2}
    \end{subfigure}
    \hfill
    \begin{subfigure}{0.19\textwidth}
        \includegraphics[width=\linewidth]{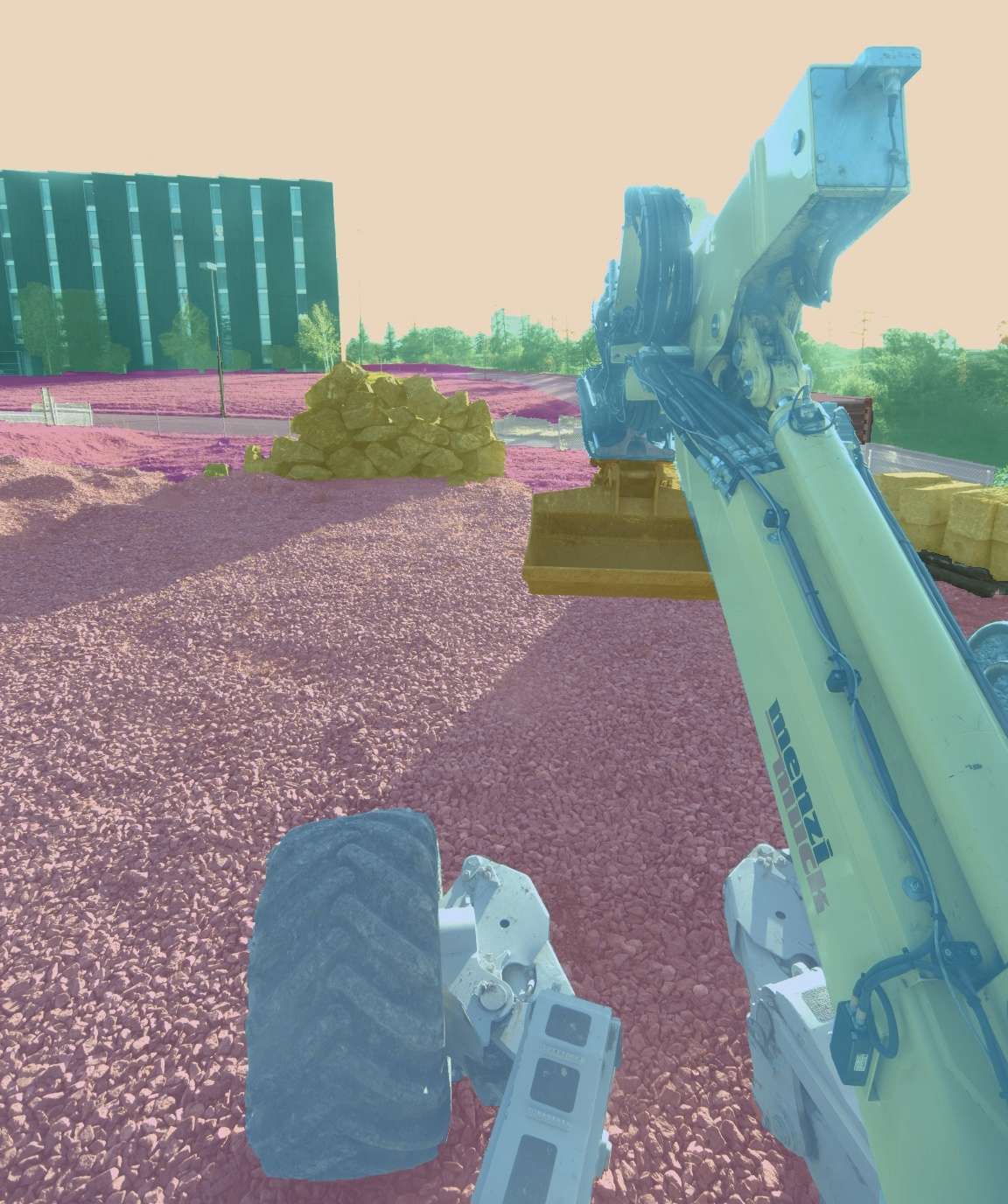}
        \label{fig:sub3}
    \end{subfigure}
    \hfill
    \begin{subfigure}{0.19\textwidth}
        \includegraphics[width=\linewidth]{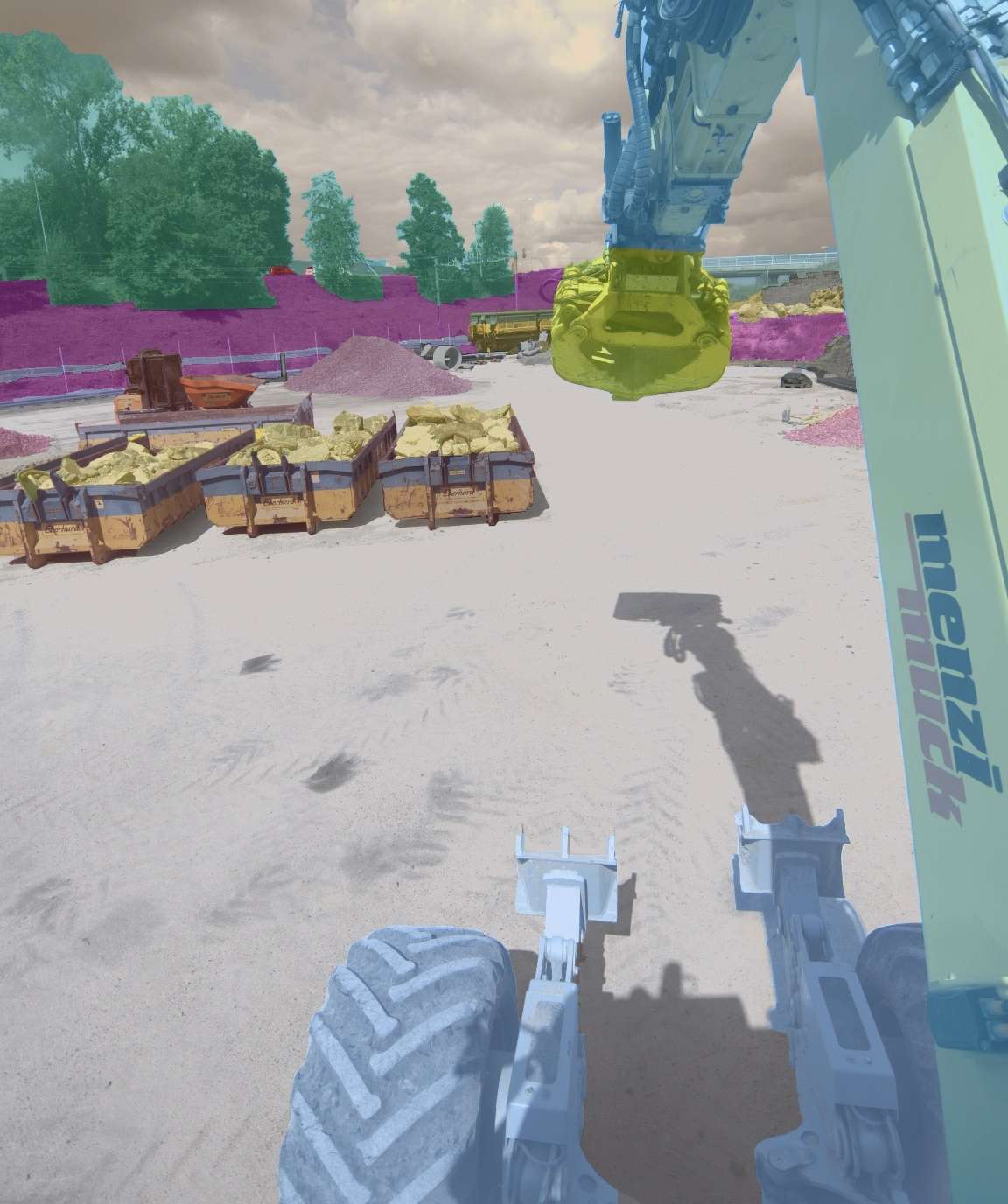}
        \label{fig:sub4}
    \end{subfigure}
    \hfill
    \begin{subfigure}{0.19\textwidth}
        \includegraphics[width=\linewidth]{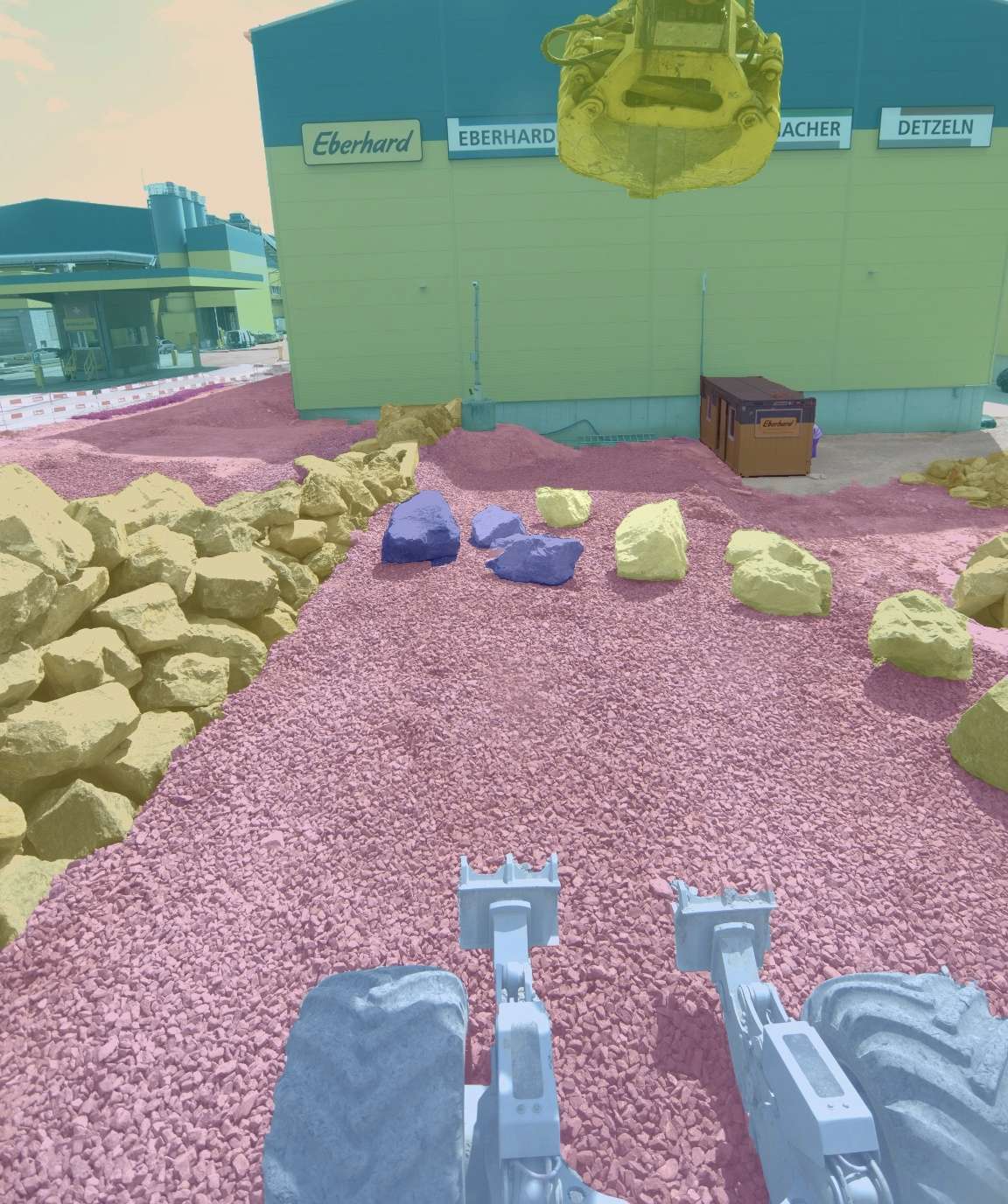}
        \label{fig:sub5}
    \end{subfigure}
    \caption{Exemplary samples from the generated and hand-labeled dataset. The 502 images published in this work are recorded in various environments, including real-world construction sites, road navigation, and natural environments.}
    \label{fig:dataset}
\end{figure*}
%%%%%%%%%%%%%%%%%%%%%%%%%%%%%%%%%%%%%%%%%%%%%%%%%%%%%%%%%%%%%%%%%

\subsection{Panoptic Segmentation}
\label{subsec:approach_panoptic_segmentation}
This study investigates two transformer-based segmentation models: Mask2Former~\cite{cheng2021mask2former} and DETR~\cite{carion2020end}. Mask2Former is an encoder-decoder transformer that can use Swin Transformer (available in \textit{tiny}, \textit{big}, and \textit{large} sizes) as a backbone encoder. On the other hand, DETR takes a different path by using attention mechanisms only during its decoding stage while still using a ResNet-50 as an encoder. We kept the original design of both models but tweaked the final layers to match our specific needs for labels.

\subsubsection{Training Procedure}
We started training with pre-existing weights from the Imagenet 21k/1k for the encoders (backbone). Moreover, we used weights from the COCO dataset training for the segmentation heads/decoder, as COCO covers a wide range of objects found in various scenes, such as roads and indoors. 
Since our work focuses on construction sites, we simplified our labels by removing unrelated ones (like food and animals). We combined similar ones into broader categories (e.g., \textit{building-other-merged}). We also added new labels important for our work, such as \textit{bucket}, \textit{gripper}, and \textit{self-arm}, refining our label list to \textbf{34} categories.

For DETR, we followed a three-step training approach: \textit{i)} first, we fine-tuned the model with our data, \textit{ii)} then fixed the main part of the model while training the segmentation parts and \textit{iii)} finally trained the whole model together for up to 200 epochs. We kept the Mask2Former training close to the original settings~\cite{cheng2021mask2former} but made some minor adjustments to fit our training on a standard GPU and to adjust the final layer for our dataset's categories. All implementational (training) details can be found in our public repository.\textsuperscript{\ref{footnote:code}}

\subsubsection{Dataset Preparation}
We curated a specialized dataset for training and testing our panoptic segmentation approach, consisting of 502 images from various construction sites, with detailed hand-labeled segmentation masks created using \emph{Segments.ai}. This includes specific construction labels like \textit{container}, \textit{stone}, and \textit{gravel-pile}.
An overview of the different sites present in the dataset can be seen in \cref{fig:dataset}. 
Since some labels (e.g., people and cars) are underrepresented in our dataset, we also incorporated COCO images - containing at least five target classes - during model fine-tuning.

As a result, the dataset is split into a training set with 1290 COCO images and 427 construction site images, while the validation set contains 75 construction site images. The dataset is open-sourced and publicly available.\footnote{\scriptsize\url{https://segments.ai/leggedrobotics/construction_site/}}
The corresponding preprocessing code is publicly accessible with the released codebase.\footnote{\scriptsize\url{https://github.com/leggedrobotics/rsl_panoptic/tree/main/panoptic_models/panoptic_models/data}}

\subsection{Dynamic Mapping and Tracking}
\label{subsection:dynamic_mapping}

\subsubsection{Dynamic \ac{LiDAR}-based Mapping in 3D}
Our dynamic mapping and tracking system processes raw \ac{LiDAR} scans, separating them into two labeled point clouds: one representing the static environment ("stuff") and the other the potentially dynamic objects ("things"). Each dynamic object is tagged with a unique tracking ID. We don't register the scans actively on the map (e.g., done in~\cite{jelavic2022open3d}), but we feed them directly to the 2D semantic mapping module described in the next subsubsection based on the current state estimation output. We use the latest version of Graph-MSF~\cite{nubert2022graph} for state and motion estimation, fusing IMU, GNSS, and LiDAR measurements.

\begin{figure}[t]
    \centering
    \includegraphics[width=\linewidth, trim={0 1cm 0 6cm}, clip]{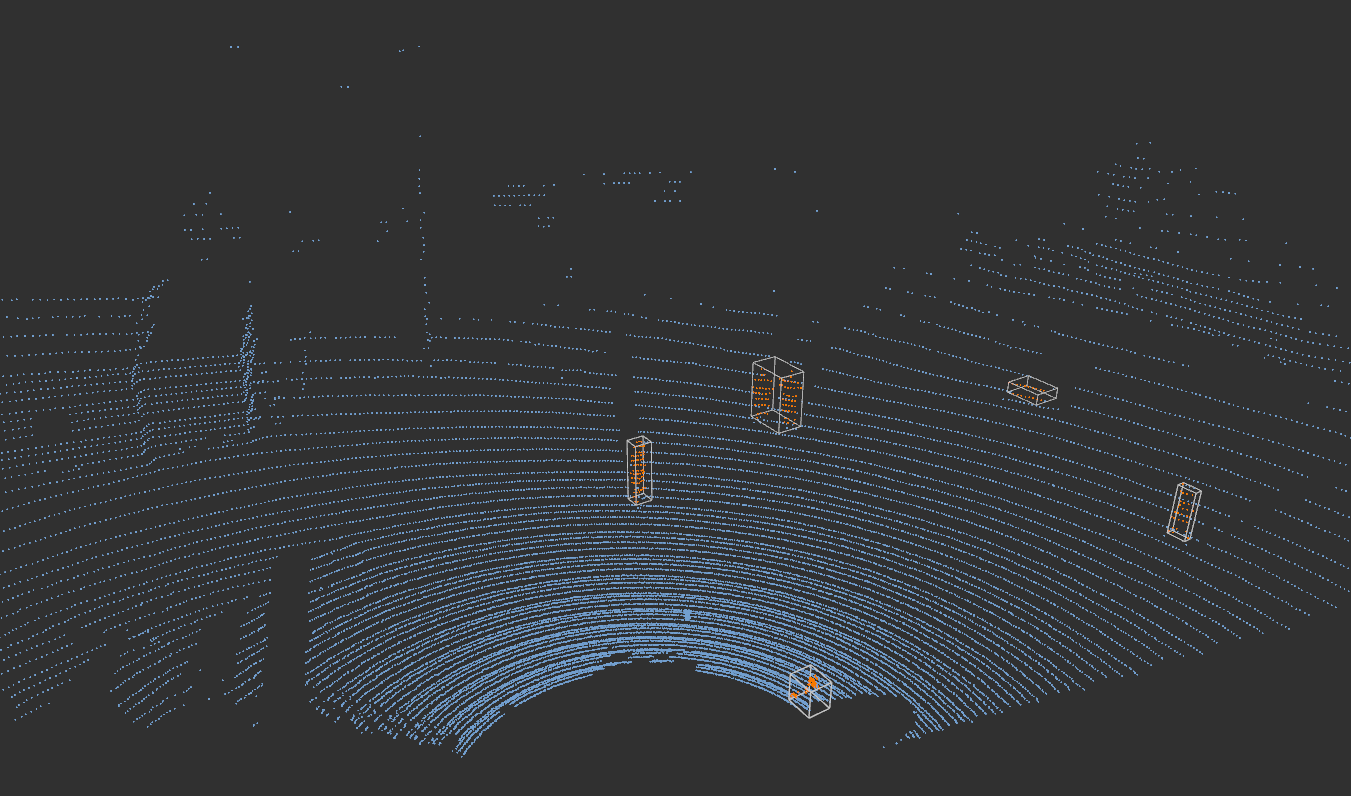}
    \caption{Segmentation of point cloud data into dynamic (red) and static (blue) elements, highlighting dynamic object tracking (e.g., people, unidentified objects) as bounding boxes.}
    \label{fig:examples_pcd}
\end{figure}

First, all \ac{LiDAR} points are projected onto the camera's image plane, labeled based on the closest pixel in the image segmentation mask. The points outside the camera's view are assigned an "unknown" label. We then use DBSCAN~\cite{ester1996dbscan} to detect clusters of points with the same label. The ground plane of the \ac{LiDAR} scan is removed for this step to get disjoint clusters. A majority vote over the constituting points' labels determines each cluster's overall label. The majority label must cross a defined threshold to minimize false detections. The cluster is assigned the "unknown" label if the label cannot be conclusively determined. The bounding boxes' position and velocity of the identified clusters are then tracked by a Kalman Filter, assuming constant velocity of the point cluster, even when leaving the field of view by retaining the classes of previous camera-based observations. All points within bounding boxes belonging to either "things" or unknown labels are added to the \textit{dynamic} cloud.
All the remaining points, including the ones within bounding boxes belonging to "stuff" labels, are assigned the \textit{static} label. 
An exemplary static and dynamic point cloud is shown in \cref{fig:examples_pcd}.

\subsubsection{Semantic Map Creation}
\label{subsubsec:semantic_map}
A semantic 2D map developed with the GridMap library~\cite{Fankhauser2016GridMapLibrary} is at the core of our mapping and tracking method. This map is divided into two primary layers, \textit{i)} \textit{static} and \textit{ii)} \textit{dynamic}, to differentiate between the non-movable surroundings ("stuff") and the movable object detections ("things") from before. Two grid-map layers are necessary to remember the surroundings, even in the presence of dynamic objects, instead of being overwritten.

The static layers use the static point cloud generated by the tracking system, while the movable one uses the tracked dynamic point cloud. Fixed parts of the environment, such as walls, terrain, and permanent structures, are categorized as non-movable, and their information in the static layer is retained, even for parts no longer visible to the mapping system. Conversely, movable objects are dynamically updated and removed as new information becomes available from the dynamic cloud.
In each update cycle, the layers merge into a single semantic layer, with \textit{dynamic} objects superseding \textit{static} map entries. For instance, if a person walks over gravel, the gravel cell's classification changes to "person" before returning to "gravel" if the person continues. 

We then produce two new maps from this merged layer: an \textit{occupancy map} based on class metadata and a \textit{cost map}. 
Semantic terrain classes like \textit{road}, \textit{gravel}, and \textit{grass} are marked as traversable, while classes like \textit{person} and \textit{fence} are non-traversable. The occupancy map aids the sampling planner in eliminating unfeasible routes, and the cost map helps to calculate trajectory costs, factoring in terrain difficulties such as mud or water, which are challenging for construction machinery. The traversability costs are manually assigned and stored in the class metadata.

Moreover, an existing site's geometric or a priori traversability map can be integrated with the occupancy map derived from the semantic map. This combination, executed with a "union" operation, is helpful for pre-known site dimensions or geofencing the machinery.

\subsection{Motion Planning}
\label{subsec:approach_navigation}
Our approach for path finding employs online RRT* implemented using the Open Motion Planning Library (OMPL). The planner checks the path's validity through an occupancy map, considering a path valid only if the occupancy values under the excavator's footprint are zero, indicating an absence of obstacles. The total trajectory cost, \(C_{\text{total}}\), is calculated as 
\begin{equation}
    C_{\text{total}} = \lambda_1 C_{\text{length}} + \lambda_2 C_{\text{semantic}}, 
\end{equation}
where \(C_{\text{length}}\) is the path's total length in meters, and \(C_{\text{semantic}}\) represents the aggregated semantic cost from the cost map. The coefficients \(\lambda_1 = 1\) and \(\lambda_2 = 0.1\) are chosen to weigh the path length and semantic costs, respectively. The semantic cost for each path segment is computed by averaging the traversability costs within the excavator's footprint, summing these averages for the total path semantic cost.

The path planner functions in real-time, consistently operating at 1 Hz with a maximum planning time of 0.95 seconds. This allows the robot to adjust its route dynamically, always seeking the most efficient path. Trajectories are calculated based on a constant velocity model from the robot's projected future position at the end of the planning cycle. If a new trajectory offers a lower cost than the current path's remaining cost, the system switches to the latest trajectory. Trajectories that pose a collision risk are assigned an infinite cost, causing the robot to stop if a potential collision is detected within a 3-meter radius until a safe path can be found.

\section{Experimental Results}\label{sec:evaluation}

We evaluate our system through both offline and online experiments. \textit{i)} Offline, we perform segmentation evaluations on our dataset (\cref{sec:exp_segmentation}). \textit{ii)} Online, we conduct system tests—including segmentation, tracking, and planning—by deploying the system on a \textit{Menzi Muck M545} excavator (\cref{sec:exp_planning}). This machine is equipped with a vertically mounted 4K \textit{Ximea xiX} PCIe camera and an \textit{Ouster OS0} \ac{LiDAR}. % We utilize the optimization-based state estimation and localization module presented in\cite{nubert2022graph} for online tracking control and map building. 
For detailed information on the robot, refer to~\cite{jud2021heap}.

\subsection{Panoptic Segmentation}
\label{sec:exp_segmentation}
In the following, we report three metrics: \ac{PQ}, \ac{SQ}, and \ac{RQ}. PQ measures overall performance by evaluating the segmentation and correct identification of each object instance. SQ assesses the accuracy of the segmented shapes regardless of their classification, while the RQ evaluates the ability to classify segmented objects correctly.

\subsubsection{Model Configurations}

For DETR, we use a ResNet50 (25M parameters) backbone pre-trained on ImageNet 1k. For Mask2Former, we train three variants, each equipped with a different-sized backbone: Swin-Tiny (29M), Swin-Big (88M), and Swin-Large (200M), all initially pre-trained on ImageNet 21k~\cite{deng2009imagenet}.

Initially, we employ DETR to determine the optimal training configuration for the best downstream performance. Our findings indicate that a three-stage training process—first focusing on the box detector, followed by the segmentation head, and finally fine-tuning the entire network—produces better outcomes than merely completing the first two stages without adjusting the backbone. This improvement is quantified in \cref{tab:detr_training}, suggesting that fine-tuning the backbone is beneficial, especially for our robot's first-person perspective dataset, which features unique objects and surfaces not widely represented in datasets like ImageNet 21k and COCO.

\begin{table}[t]
    \renewcommand{\arraystretch}{1.3}
    \caption{Evaluation of the two DETR training methods.}
    \label{tab:detr_training}
    \centering
    \begin{tabular}{c|ccc}
        \hline
        \textbf{Method} & \textbf{PQ} & \textbf{SQ} & \textbf{RQ} \\
        \hline
        Freeze backbone & 0.35 & 0.61 & 0.44 \\
        Fine-tune backbone & \textbf{0.41} & \textbf{0.62} & \textbf{0.52} \\
        \hline
    \end{tabular}
\end{table}

For Mask2Former, we found the default fine-tuning procedure\cite{cheng2021mask2former} to work well in practice.

\begin{table}[b]
    \renewcommand{\arraystretch}{1.3}
    \caption{Evaluation results of DETR and Mask2Former (M2F) on the "construction site" validation set.}
    \label{tab:segmentation_results}
    \centering
    \begin{tabular}{c|cccc}
        \hline
        \textbf{Model} & \textbf{PQ} & \textbf{SQ} & \textbf{RQ} & \textbf{Images/s} \\
        \hline
        DETR & 0.41 & 0.62 & 0.52 & \textbf{12} \\
        M2F Swin-Tiny & \textbf{0.68} & \textbf{0.79} & \textbf{0.81} & 8 \\
        M2F Swin-Big & 0.63 & 0.77 & 0.73 & 3 \\
        M2F Swin-Large & \textbf{0.68} & \textbf{0.79} & 0.80 & 1 \\
        \hline
    \end{tabular}
\end{table}

\subsubsection{Performance Comparison}

Our results, summarized in \cref{tab:segmentation_results}, highlight the superior performance of the fully transformer-based Mask2Former models. All models are trained until convergence. Due to memory constraints on the training GPU (\textit{Nvidia RTX 3090}), the Swin-Big and Swin-Large models were fine-tuned with a batch size of only 1.

\begin{figure}[t]
    \centering
    \begin{subfigure}{0.5\linewidth}
        \centering
        \includegraphics[width=.95\linewidth]{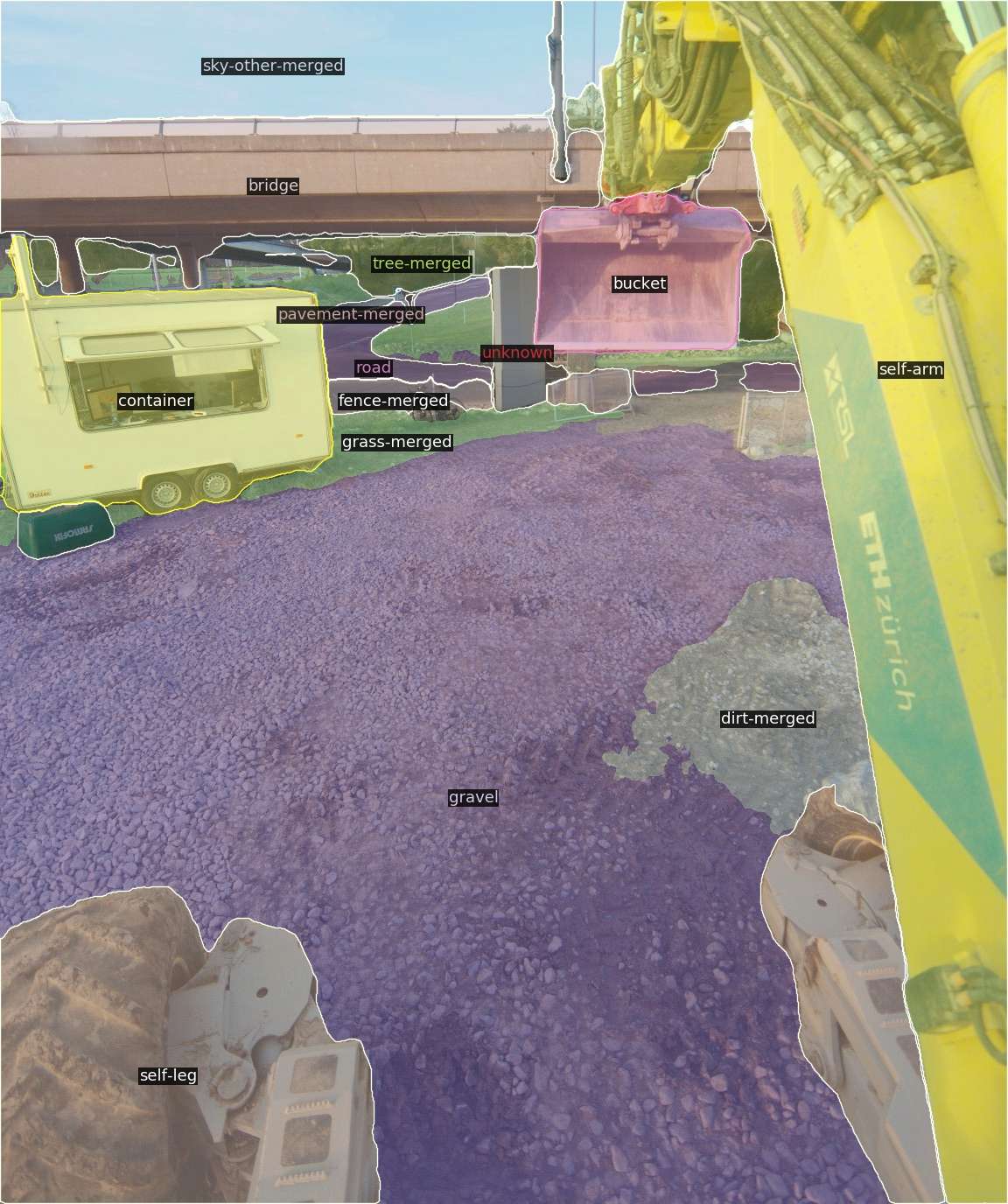}
        \label{subfig:segmentation_1}
    \end{subfigure}%
    \begin{subfigure}{0.5\linewidth}
        \centering
        \includegraphics[width=.95\linewidth]{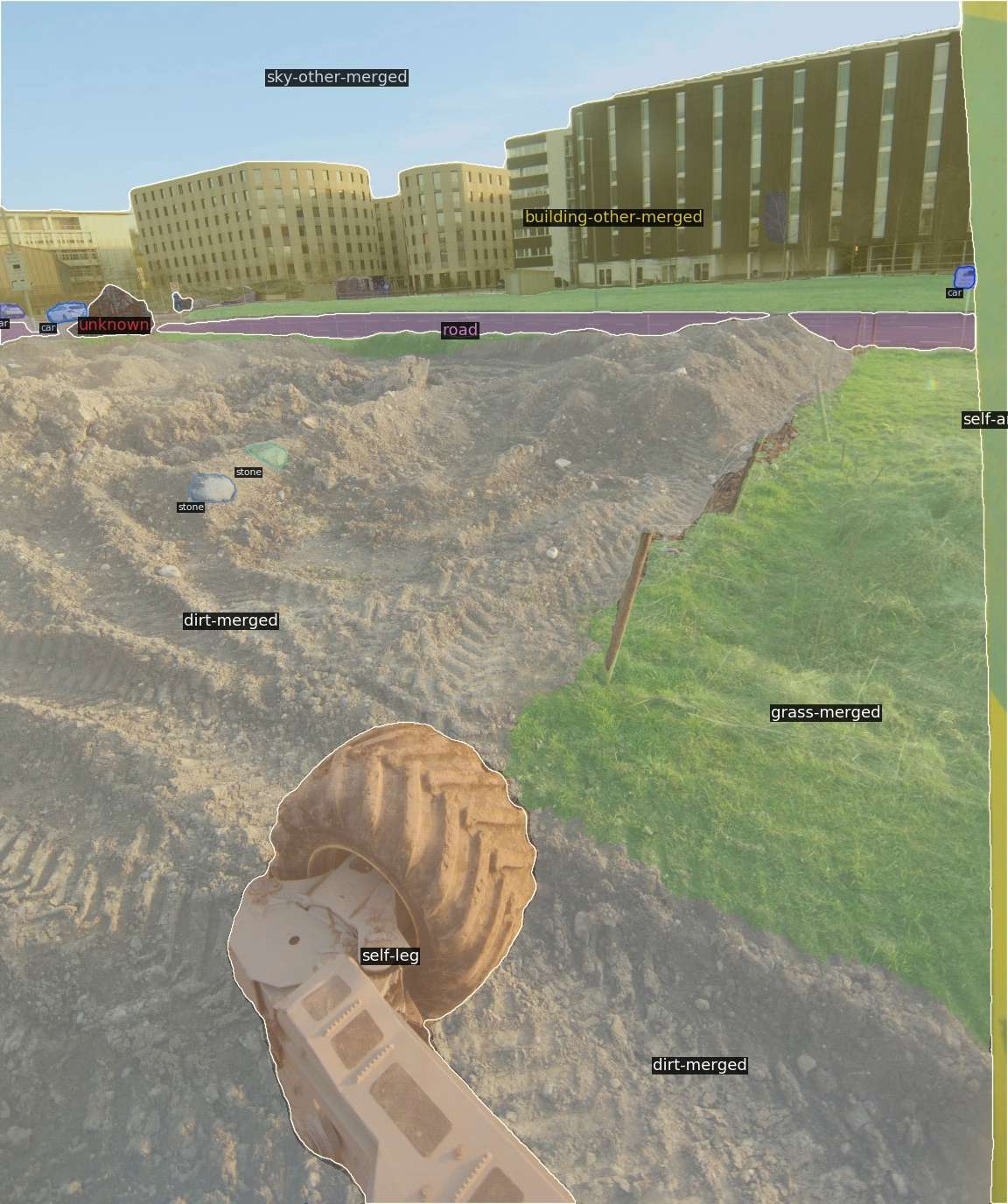}
        \label{subfig:segmentation_2}
    \end{subfigure}
    \caption{Two exemplary segmentation masks produced by Mask2Former.}
    \label{fig:example_segmentations}
\end{figure}

Despite its smaller size, the Swin-Tiny backbone performs well, suggesting that the larger models may require more extensive data and computational resources for optimal training. Additionally, Mask2Former with Swin-Tiny exhibits favorable inference throughput rates, making it an ideal candidate for real-time applications on our robot equipped with an \textit{Nvidia RTX 3080Ti}. \Cref{fig:example_segmentations} shows two exemplary segmentation estimates.

\begin{figure}[b]
    \centering
    \begin{subfigure}{0.5\linewidth}
        \centering
        \includegraphics[width=.95\linewidth]{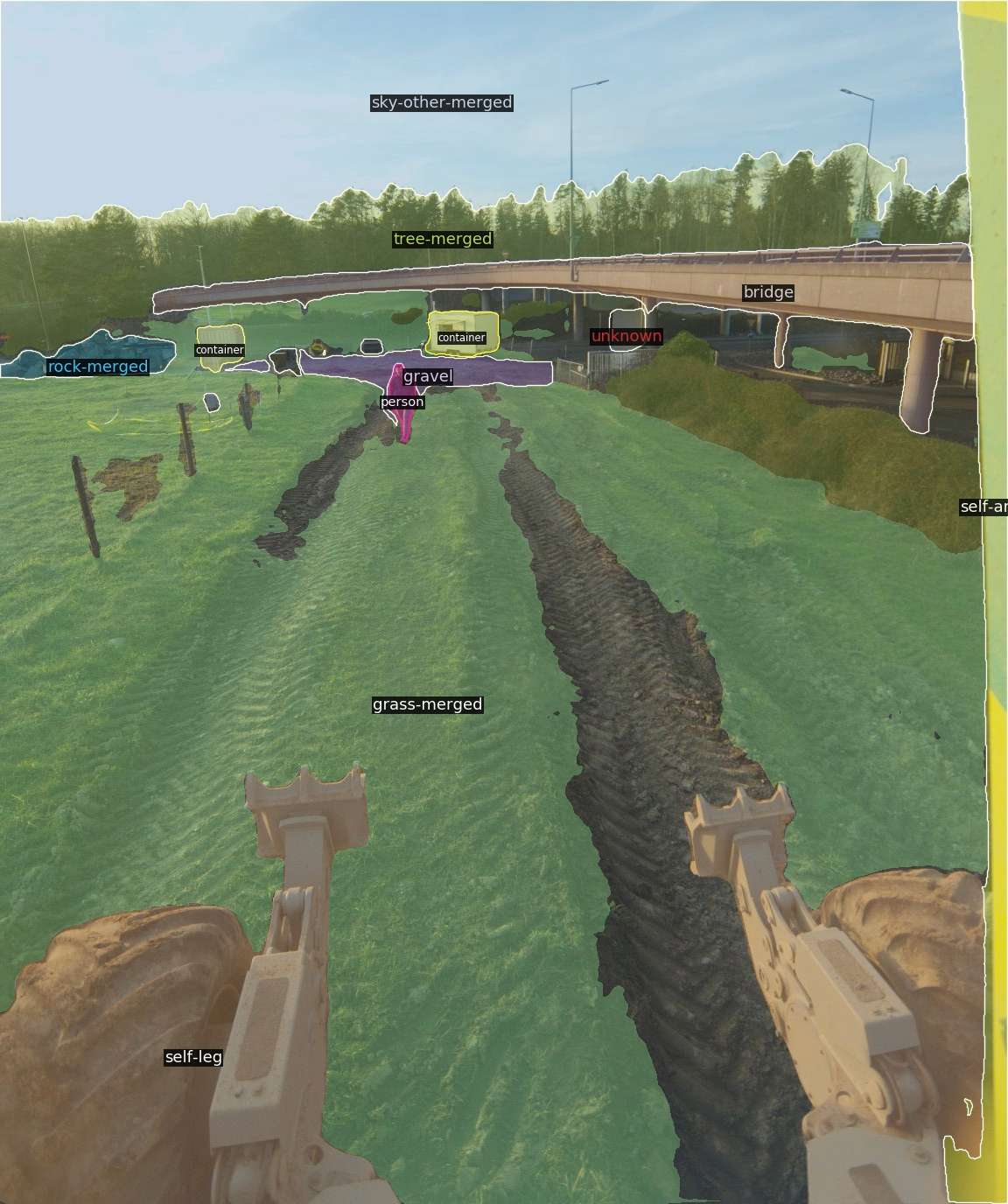}
        \label{subfig:threshold_problem_1}
    \end{subfigure}%
    \begin{subfigure}{0.5\linewidth}
        \centering
        \includegraphics[width=.95\linewidth]{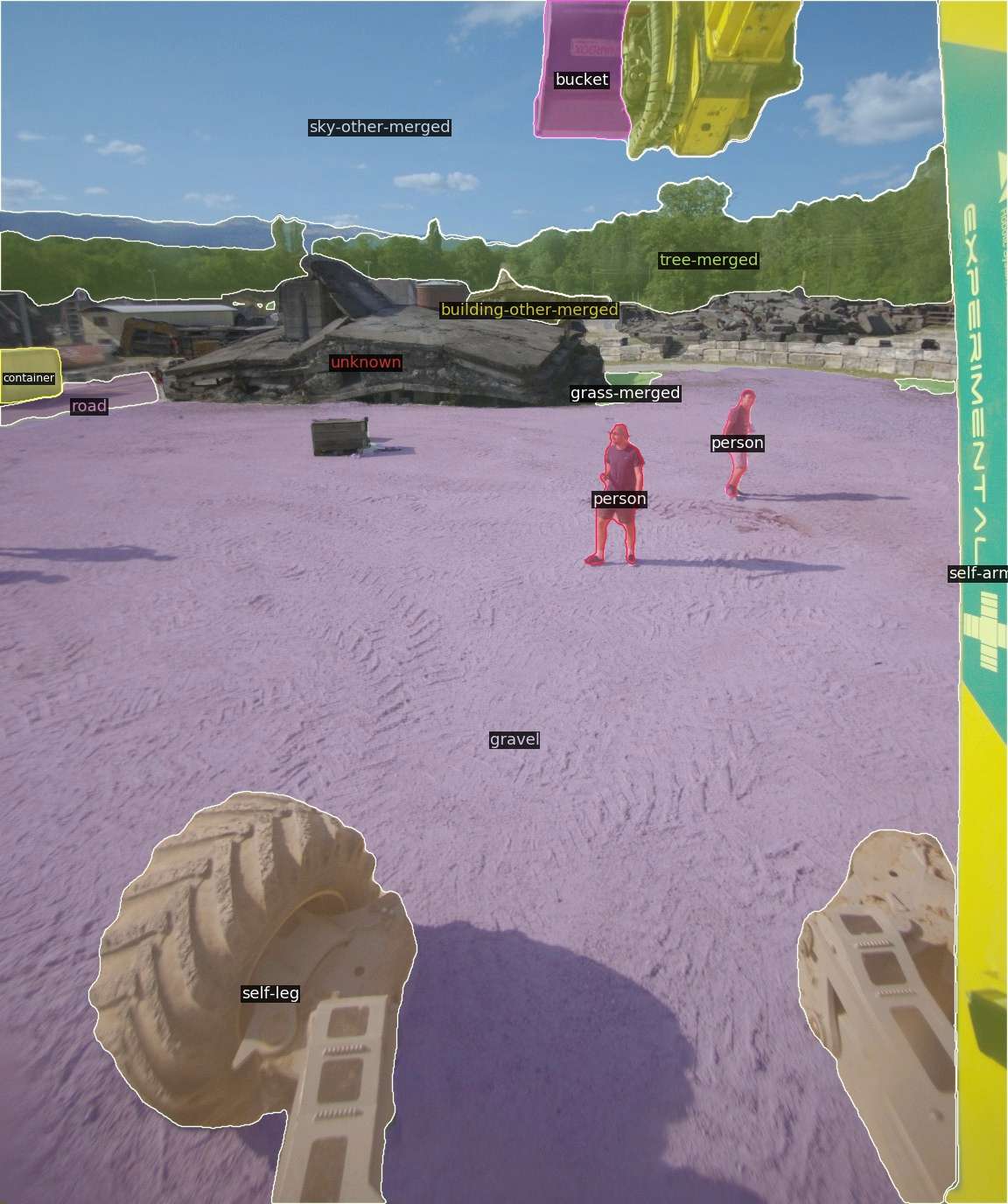}
        \label{subfig:threshold_problem_2}
    \end{subfigure}
    \caption{Problematic segmentation due to confidence threshold: Mask2Former's inability to decisively segment ambiguous scenes such as dirt tracks and potentially misclassified debris.}
    \label{fig:confidence_threshold_problem}
    \vspace{-0.5cm}
\end{figure}

A critical aspect of all segmentation systems is setting the suitable confidence threshold, which determines whether a prediction is certain enough to be considered valid. We observed that Mask2Former often struggles to classify terrain types like dirt, gravel, or pavement, which could plausibly belong to multiple categories. This occurs because the model distributes its confidence across several classes, resulting in no single prediction meeting the confidence threshold. This challenge is illustrated in \cref{fig:confidence_threshold_problem}, highlighting a limitation inherent to current vision systems with a fixed set of categories.

\subsubsection{Dataset Evaluation}

Next, we examine how the distribution of classes in the training dataset impacts the performance of downstream tasks. Given our dataset's modest size of around 500 images, we investigated whether models trained solely on our data would overfit, diminishing their effectiveness in segmenting unfamiliar scenes. By training DETR on either solely our dataset or a combination with a subset of the COCO dataset, we observed a significant performance degradation when using only our dataset, as shown in \cref{tab:dataset_evaluation}. This outcome points to severe overfitting. We discovered that a ratio of about 3:1 (COCO to our dataset) best improves generalization without compromising segmentation accuracy in our specific domain. Hence, we included a COCO dataset subset in all training runs. We advise incorporating parts of a larger, diverse dataset when working with small datasets to mitigate the risk of catastrophic forgetting.

\begin{table}[t]
    \renewcommand{\arraystretch}{1.3}
    \caption{DETR performance on construction site and COCO-only validation sets, with different training sets.}
    \label{tab:dataset_evaluation}
    \centering
    \begin{tabular}{c|c|ccc}
        \hline
        \textbf{Train set} & \textbf{Val set} & \textbf{PQ} & \textbf{SQ} & \textbf{RQ} \\
        \hline
        \multirow{2}{*}{Custom} & COCO val & 0.06 & 0.18 & 0.09 \\
                                & Custom val & 0.41 & 0.54 & 0.51 \\
        \hline
        \multirow{2}{*}{Custom + COCO} & COCO val & 0.26 & 0.46 & 0.36 \\
                                       & Custom val & 0.41 & 0.53 & 0.52 \\
        \hline
    \end{tabular}
\end{table}

Another interesting aspect is determining the optimal size of the dataset. The necessary volume of data depends on the application's tolerance for false positives and negatives. For navigation tasks, errors can dramatically compromise system reliability (triggering unnecessary replanning or stops) or safety (overlooking obstacles).

\Cref{tab:dataset_size} displays how Mask2Former's performance scales with training dataset size on our validation set. The model's PQ and RQ show near-linear improvement with increasing training images without showing signs of plateauing even at approximately 400 images, indicating potential benefits from an even larger dataset. On the other hand, the insensitivity of SQ to dataset size in this context can be attributed to the pre-trained model's already developed competence in capturing the geometric properties of objects, which generalizes well across different object categories.

\begin{table}[b]
    \centering
    \caption{Performance of the Mask2Former Swin-Tiny model on panoptic segmentation for different "construction site" dataset fractions.}
    \label{tab:dataset_size}
    \begin{tabular}{c|ccccc}
        \hline
        \textbf{Fraction} & \textbf{20\%} & \textbf{40\%} & \textbf{60\%} & \textbf{80\%} & \textbf{100\%} \\
        \hline
        \textbf{PQ} & 0.53 & 0.57 & 0.59 & 0.62 & \textbf{0.68} \\
        \textbf{RQ} & 0.63 & 0.67 & 0.70 & 0.72 & \textbf{0.81} \\
        \textbf{SQ} & 0.77 & 0.77 & 0.78 & \textbf{0.79} & 0.78 \\
        \hline
    \end{tabular}
    \vspace{-0.5cm}
\end{table}

\subsection{Motion Planning with Semantic Understanding}
\label{sec:exp_planning}
We conducted autonomous navigation trials in a controlled testing field and at the Rescue Troop Training Center in Avully, Switzerland. In the testing field, the system's performance was evaluated in narrow, obstacle-dense environments containing challenging objects like fences, poles, and buckets. These obstacles, difficult to distinguish geometrically or using \ac{LiDAR}, provided a rigorous test of the system's perception capabilities.

At Avully, we assessed navigation in open spaces, focusing on road-following and dynamic obstacle handling, such as emergency stops or rerouting around people. The supplementary video provides a demonstration of the field deployment.

During all tests, Mask2Former and the \ac{LiDAR}-based obstacle detection system performed without critical misclassifications. The online planner consistently found reliable solutions. Although occasional attempts at pathfinding would fail due to dynamic changes, the planner's continuous re-evaluation ensured robust navigation suitable for long-duration operations, as shown in~\cite{terenziAutonomousExcavationPlanning2023}.

\begin{figure*}[t]
    \centering
    % First row
    % \begin{subfigure}[b]{0.32\textwidth}
        \includegraphics[width=0.32\linewidth, clip, trim=164px 0px 165px 0px]{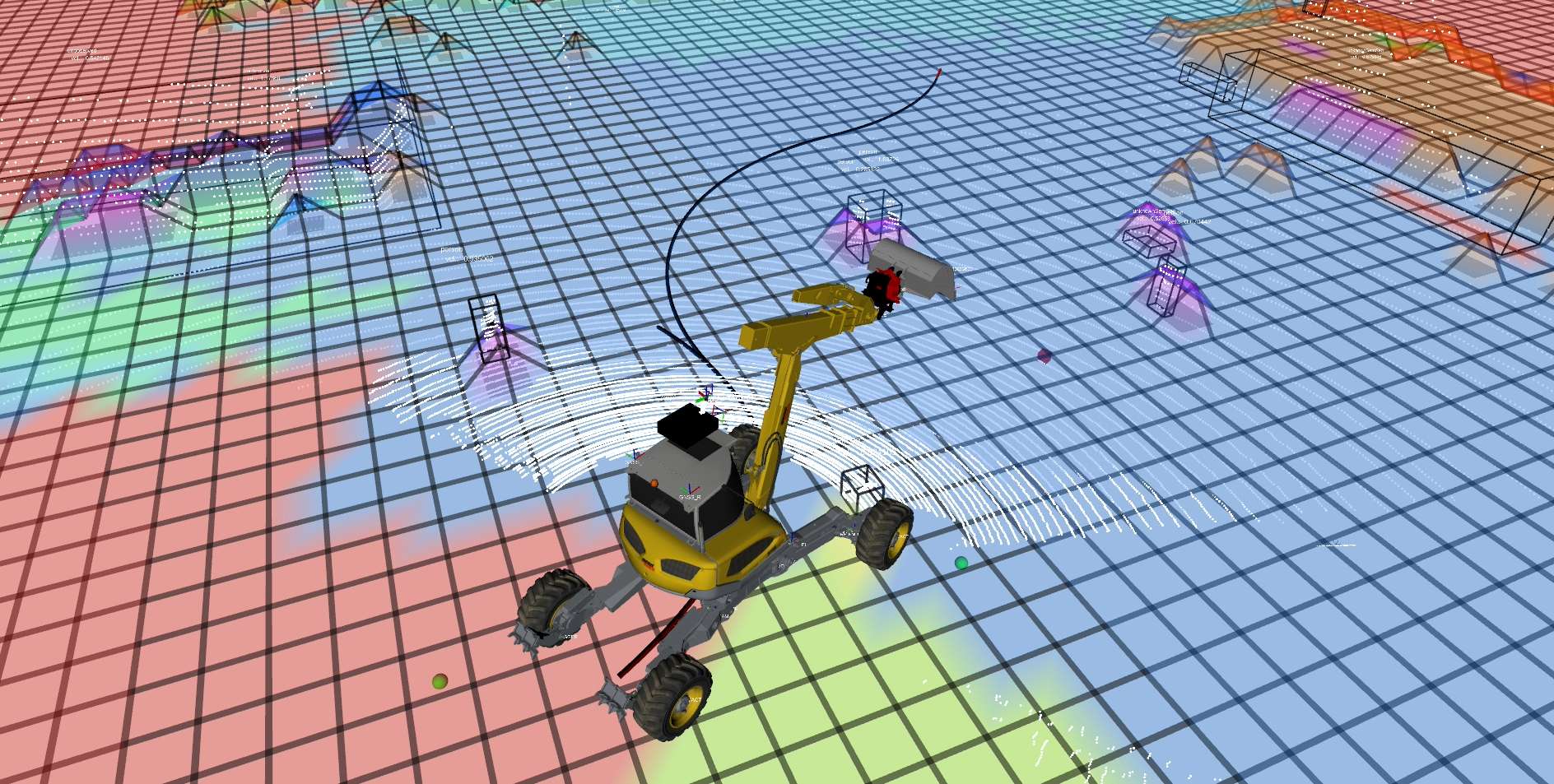}
        % \caption{}
        \label{fig:first_row_first_image}
    % \end{subfigure}
    \hfill % Use \hfill to add horizontal space between the figures
    % \begin{subfigure}[b]{0.32\textwidth}
        \includegraphics[width=0.32\linewidth, clip, trim=0px 6px 0px 6px]{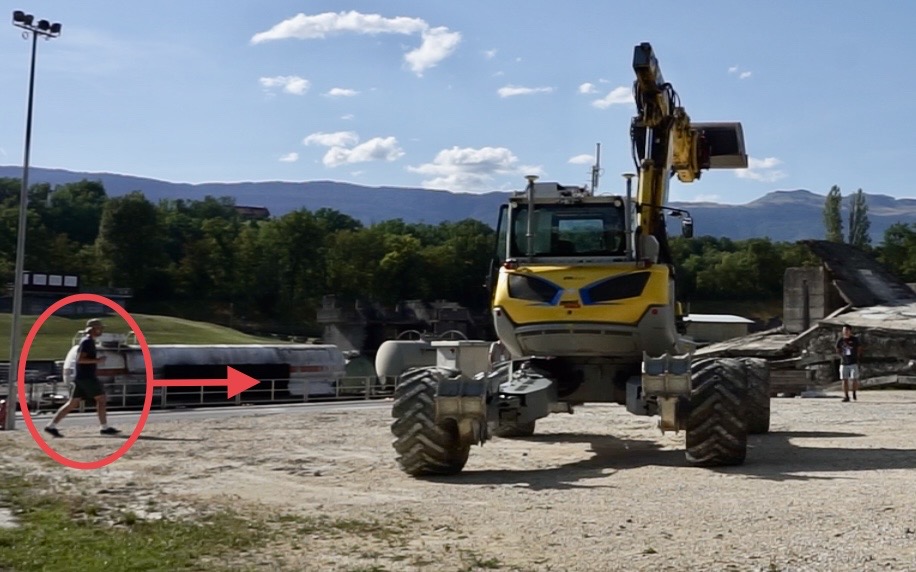}
        % \caption{}
        \label{fig:first_row_second_image}
    % \end{subfigure}
    \hfill % Use \hfill to add horizontal space between the figures
    % \begin{subfigure}[b]{0.32\textwidth}
        \includegraphics[width=0.32\linewidth,  clip, trim=0px 336px 0px 336px]{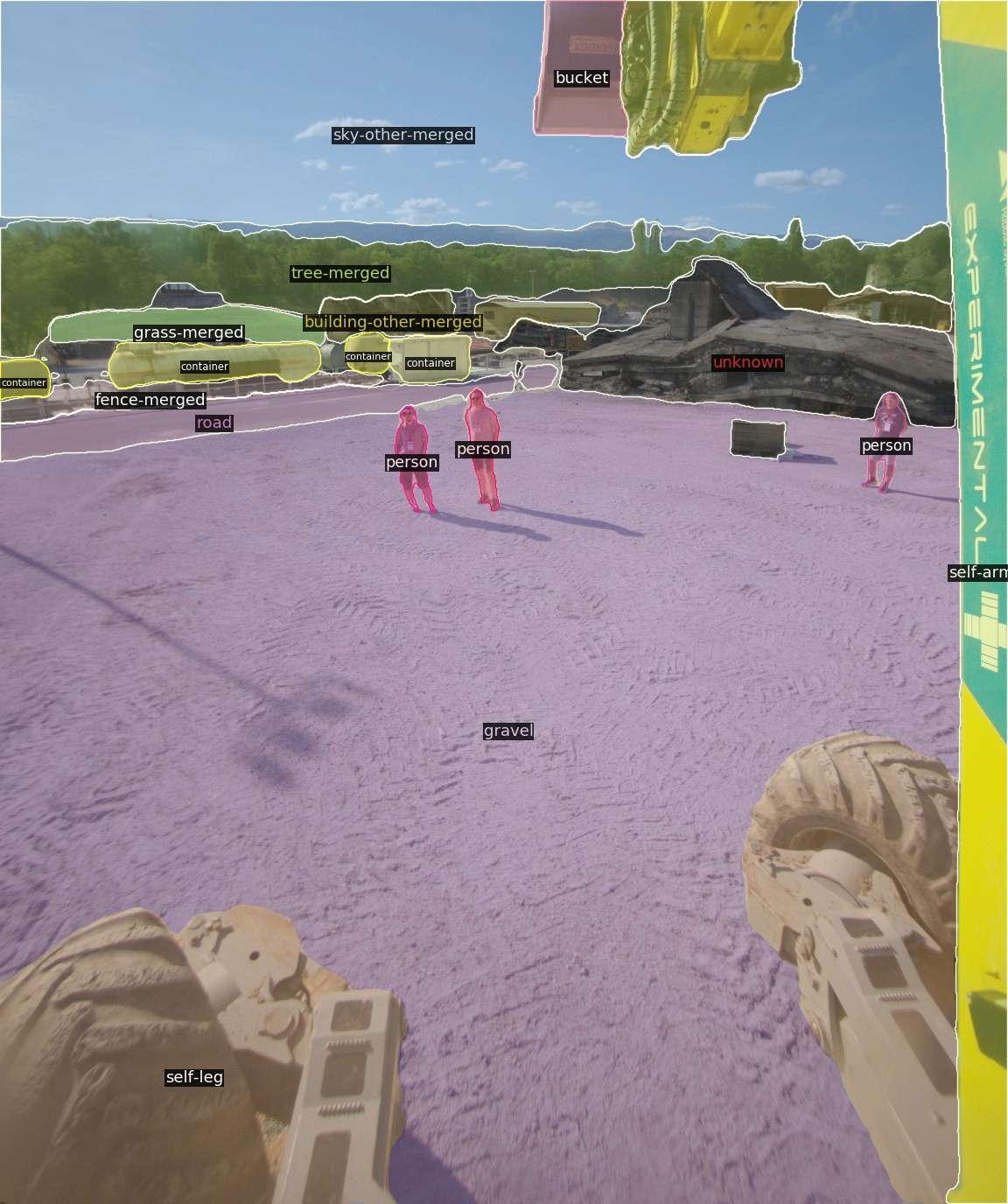}
        % \caption{}
        \label{fig:first_row_third_image}
    % \end{subfigure}
    
    % Add some vertical space between the rows
    \vspace{5pt}
    
    % Second row (Repeat with different images or the same for placeholder)
    % \begin{subfigure}[b]{0.32\textwidth}
        \includegraphics[width=0.32\linewidth, clip, trim=164px 0px 165px 0px]{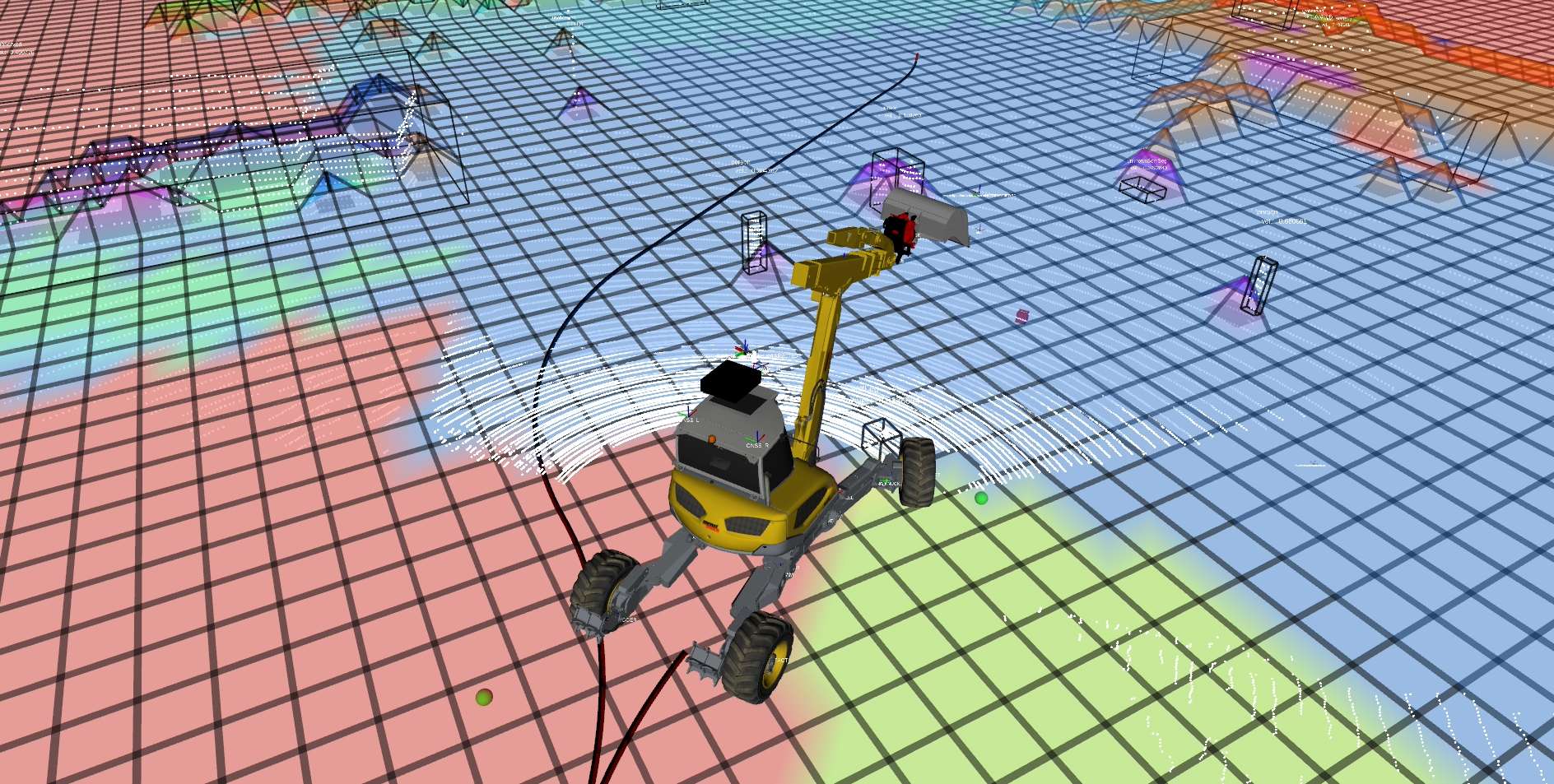}
        % \caption{}
        \label{fig:second_row_first_image}
    % \end{subfigure}
    \hfill
    % \begin{subfigure}[b]{0.32\textwidth}
        \includegraphics[width=0.32\linewidth, clip, trim=0px 0px 0px 0px]{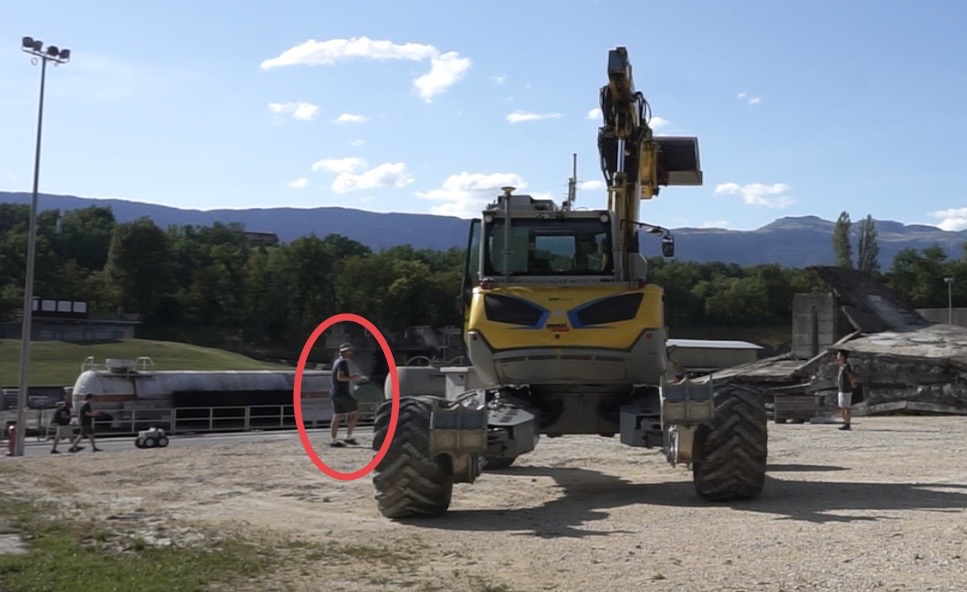} % Replace with another image for your actual content
        % \caption{}
        \label{fig:second_row_second_image}
    % \end{subfigure}
    \hfill
    % \begin{subfigure}[b]{0.32\textwidth}
        \includegraphics[width=0.32\linewidth, clip, trim=0px 336px 0px 336px]{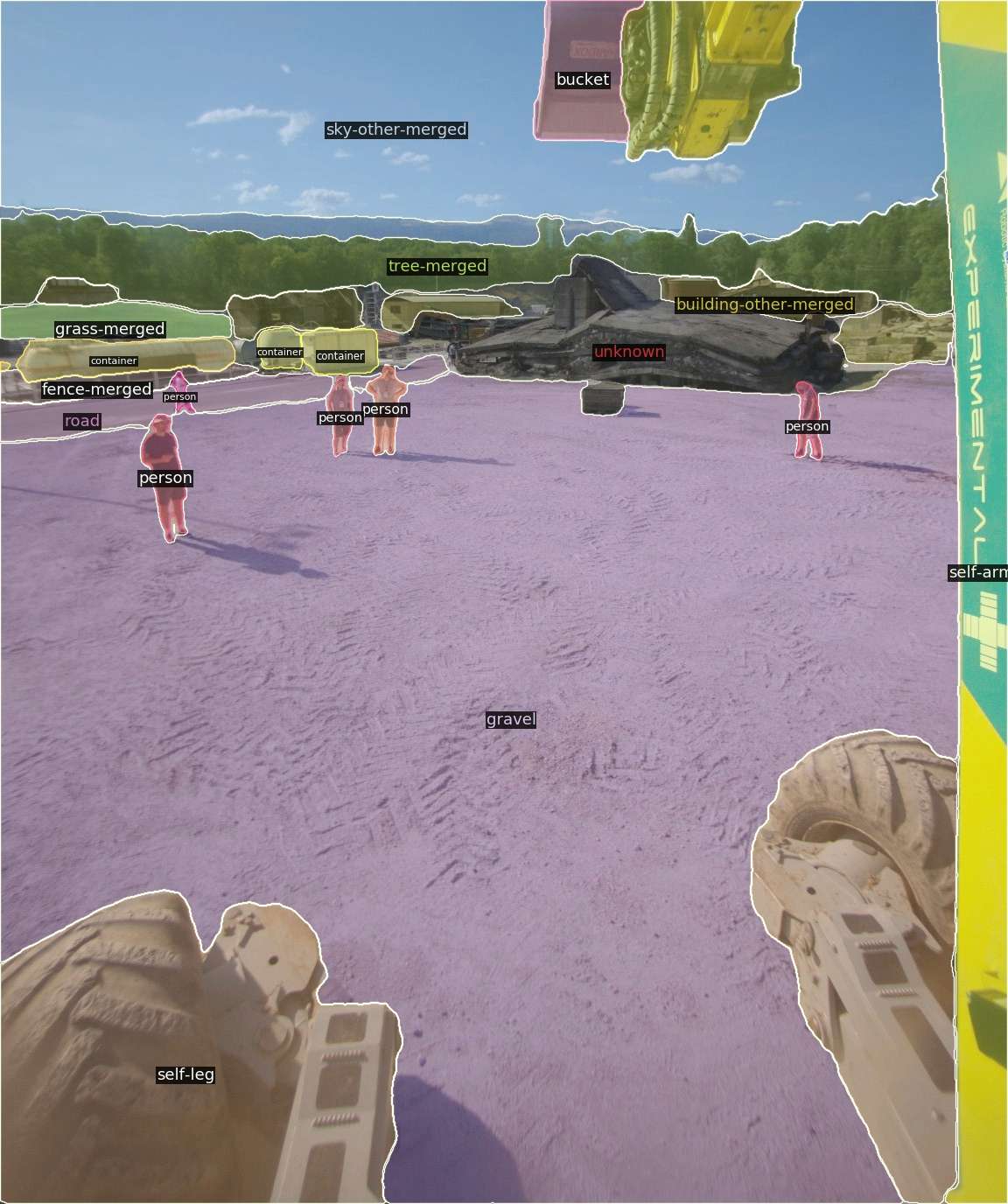} % Replace with another image for your actual content
        % \caption{}
        \label{fig:second_row_third_image}
    % \end{subfigure}
    \caption{Two consecutive snapshots from an "adversarial test" of the navigation system conducted in Avully, where humans attempt to obstruct the robot by crossing its planned paths. %This scenario involves four individuals blocking the path of an excavator. 
    The first column shows the system's semantic mapping and object tracking, with %bounding boxes highlighting dynamic obstacles and 
    different colors for the semantic classes (untraversable areas elevated). The planned path is illustrated in black. %The second column presents a third-person view of the scene, while t
    The third column offers a segmented, first-person view. The online RRT* planner dynamically generates alternative routes.}
    \label{fig:full_page_image_layout}
    \vspace{-0.5cm}
\end{figure*}

\Cref{fig:full_page_image_layout} shows snapshots from an "adversarial test" during our field deployment in Avully, where the navigation system recalculated a new plan when dynamic obstacles (people) obstructed the initial path.

\section{Conclusions \& Future Work}\label{sec:conclusions}

This work introduced a comprehensive panoptic scene understanding system tailored for construction environments. We presented a fine-tuned Mask2Former-based segmentation model that integrates visual and geometric data to generate dynamic and static semantic maps. The system tracks objects over time, maintaining their semantic identities outside the camera’s field of view and effectively segmenting the environment into static and dynamic layers.

We demonstrated the practical use of this system by applying it to a navigation task. The generated panoptic maps were utilized in an online RRT* planner for robust, real-time path planning in dynamic environments. While the navigation task served as a case study, the panoptic segmentation system is generalizable and can be applied to other robotics and perception tasks.

Future improvements include addressing ambiguities in classifying similar terrain by incorporating open-vocabulary systems and enhancing point cloud detection with \ac{NN}-based methods. Extending the 2D semantic map to a full 3D representation would enable navigation under complex structures. The code and dataset are publicly available to facilitate further research and application in this domain.

%%%%%%%%%%%%%%%%%%%%%%%%%%%%%%%%%%%%%%%%%%%%%%%%%%%%%%%%%%%%%%%%%%%%%%%%%%%%%%%%

%%%%%%%%%%%%%%%%%%%%%%%%%%%%%%%%%%%%%%%%%%%%%%%%%%%%%%%%%%%%%%%%%%%%%%%%%%%%%%%%
                                                                                                                                                                                                                                                                                                                                                                                                                                                                                                                                                                                                                                                                                                                                                                                                                                                                                                                                                                                                                                                                                                                                                                                                                                                                                                                                                                                                                                                                                                                                                                                                                                                                                                                                                                                                                                                                                                                                                                                                                                                                                                                                                                                                                                                                                                                                                                                                                                                                                                                                                                                                                                                                                                                                                                                                                                                                                                                                                                                                                                                                                                                                                                                                                                                                                                                                                                                                                                                                                                                                                                                                                                                                                                                                                                                                                                                                                                                                                                                                                                                                                                                                                        
\bibliographystyle{styles/acra}
\bibliography{references}

\end{document}